\documentclass[conference,a4paper]{IEEEtran}

\makeatletter
\newcommand{\linebreakand}{%
  \end{@IEEEauthorhalign}
  \hfill\mbox{}\par
  \mbox{}\hfill\begin{@IEEEauthorhalign}
}
\makeatother

\IEEEoverridecommandlockouts

\usepackage{cite}
\usepackage{amsmath,amssymb,amsfonts}
\usepackage{algorithmic}
\usepackage{graphicx}
\usepackage{textcomp}
\usepackage{xcolor}
\usepackage{booktabs}
\usepackage{multirow}
\usepackage{array}
\usepackage{flafter}
\usepackage{placeins}
\usepackage[
    a4paper,
    top=19.1mm,
    bottom=19.1mm,
    left=14.3mm,
    right=14.3mm
]{geometry}

\newcommand{\supercite}[1]{\textsuperscript{\cite{#1}}}

\def\BibTeX{{\rm B\kern-.05em{\sc i\kern-.025em b}\kern-.08em
    T\kern-.1667em\lower.7ex\hbox{E}\kern-.125emX}}

\begin{document}

\flushbottom

\title{IMM-based Multiple Object Tracking using a State Prediction Neural Network}

\author{
\makebox[\textwidth][c]{%
\begin{minipage}[t]{0.31\textwidth}
\centering
1\textsuperscript{st} Chan-Bin Lim\\[2pt]
\textit{Robotics Program}\\
\textit{Korea Advanced Institute of}\\
\textit{Science and Technology (KAIST)}\\
Daejeon, Republic of Korea, 34051
\end{minipage}
\hfill
\begin{minipage}[t]{0.31\textwidth}
\centering
2\textsuperscript{nd} Dong-Hee Paek\\[2pt]
\textit{Mechanical Engineering Research Institute}\\
\textit{Korea Advanced Institute of}\\
\textit{Science and Technology (KAIST)}\\
Daejeon, Republic of Korea, 34141
\end{minipage}
\hfill
\begin{minipage}[t]{0.31\textwidth}
\centering
3\textsuperscript{rd} Seung-Hyun Kong\textsuperscript{*}\\[2pt]
\textit{Cho Chun Shik Graduate School of Mobility}\\
\textit{Korea Advanced Institute of}\\
\textit{Science and Technology (KAIST)}\\
Daejeon, Republic of Korea, 34051
\end{minipage}
}
\thanks{*Corresponding author: skong@kaist.ac.kr}
}

\maketitle

\begin{abstract}
Object tracking is essential for autonomous vehicles to avoid obstacles and plan routes. Radar maintains detection performance even in adverse weather and can measure relative velocity through the Doppler effect, making it well suited for object tracking. In this paper, we propose a data-driven state PRedictor-based Interacting Multiple Model tracking method (PR-IMM) that improves nonlinear object-motion representation while preserving the stability and interpretability of physics-based motion models. The proposed method employs a Transformer-based PRediction model (PR) that incorporates radar Doppler measurements to predict object displacement. The PR model is integrated into the IMM as a mode alongside the CV, CA, and CT motion models, and their prior positions are dynamically combined according to the mode probabilities. Experimental results show that PR-IMM reduces position-estimation error by 57.3\% over the IMM and by 16.5\% over the PR, while reducing ID switches by 25.3\% and improving IDF1 by 9.6\% over the IMM.
\end{abstract}

\begin{IEEEkeywords}
Radar, multiple object tracking, Kalman filter, interacting multiple model, Transformer.
\end{IEEEkeywords}

\section{Introduction}

Object tracking detects objects from sensor data and continuously estimates their locations. With the development of autonomous driving and advanced driver assistance systems (ADAS), it has become a core technology for safe driving decisions\supercite{han2025,yoon2024,jung2025}. Cameras, LiDAR, radar, and other sensors are widely used for this purpose. Among them, radar has attracted particular attention for object tracking because it maintains robust detection performance in adverse weather such as rain and fog and can directly measure relative velocity through the Doppler effect\supercite{cho2019,kong2013,kong2014}.

Accordingly, various approaches have been proposed for radar-based object tracking, and early studies mainly relied on motion models implemented with Kalman filters (KF)\supercite{jan2013,manjunath2018}. Representative models include Constant Velocity (CV), Constant Acceleration (CA), and Constant Turn Rate (CT). Each assumes a specific motion pattern and defines a state-transition model to predict the next object position. Because a single motion model cannot flexibly represent diverse motion patterns, the Interacting Multiple Model (IMM), which dynamically combines multiple motion models according to their mode probabilities, has been widely adopted\supercite{blom1988}. However, motion-model-based approaches depend on predefined motion patterns and therefore have limited ability to represent the nonlinear object motions that occur in real road environments\supercite{lv2024}.

To overcome these limitations, recent studies have actively applied deep-learning-based trajectory prediction models to object tracking\supercite{kim2025bayes}. In particular, Transformer-based models\supercite{vaswani2017} effectively model long-range dependencies in time-series data through self-attention and have shown strong performance in predicting future positions from past trajectories\supercite{kim2024transformer}. Because such data-driven models do not assume a particular motion pattern, they can learn nonlinear motions that are difficult to describe with CV, CA, or CT models. On the other hand, their predictions can become unstable when observation noise is large or when the input lies outside the training distribution\supercite{tang2024}.

Motion-model-based and data-driven approaches therefore have complementary strengths. Data-driven models can directly learn diverse motion patterns without predefining dynamic assumptions, while the Kalman filters underlying physics-based motion models provide stable state estimation by explicitly handling measurement uncertainty through covariance. Combining these strengths is thus a promising way to improve object-tracking performance.

From this perspective, we propose a PR (PRediction model)-IMM tracker that integrates a Transformer-based prediction model into the IMM framework. The proposed method uses a data-driven prior state from a Transformer predictor learned from radar measurements together with physics-based prior states from CV, CA, and CT models. These priors are probabilistically combined through IMM mode probabilities so that the tracker can adaptively exploit the most suitable information under each motion condition, targeting robust tracking performance that cannot be achieved by a single motion model or a data-driven predictor alone.

The main contributions of this paper are as follows:
\begin{itemize}
    \item We propose a Transformer-based network that considers radar Doppler measurements for accurate state prediction.
    \item We propose a PR-IMM multiple-object-tracking framework in which a data-driven Transformer predictor and physics-based CV, CA, and CT motion models operate as IMM modes.
    \item Using the large-scale radar point-cloud dataset RadarScenes\supercite{schumann2021}, we demonstrate that the proposed method significantly improves both position-estimation accuracy and object-tracking performance over single models and a conventional IMM (CV+CA+CT).
\end{itemize}

The remainder of this paper is organized as follows. Section II reviews related work. Section III describes the proposed PR-IMM tracker in detail. Section IV presents the experimental setup and results, and Section V concludes the paper.

\section{Related Works}

This section reviews representative object-tracking methods based on Kalman filters, IMM, deep learning, and combinations of Kalman filtering and neural networks.

\subsection{Kalman-Filter-Based Object Tracking}

CV, CA, and CT motion models are among the most common choices in Kalman-filter-based tracking. SORT combines a CV-model Kalman filter with the Hungarian algorithm for real-time multiple-object tracking\supercite{bewley2016}, while CA-model Kalman filtering has also been used to estimate vehicle motion states from automotive radar measurements\supercite{li2019}. The CV model is robust for constant-speed motion but weak under abrupt acceleration. The CA model mitigates this limitation but can accumulate larger inertial errors in the lateral direction. The CT model is advantageous for turning maneuvers but may exhibit larger errors during straight constant-speed motion. Thus, a single motion model is optimized for a particular motion assumption and inevitably involves a trade-off; prediction errors can increase in real traffic where objects combine straight motion, acceleration, deceleration, and turning.

\subsection{IMM-Based Object Tracking}

IMM-based tracking commonly combines CV, CA, and CT models. Previous studies have applied IMM to 3D multiple-object tracking to reduce ID switches\supercite{li2024immekf} and to improve tracking under diverse motion patterns\supercite{liu2025immmot}. By dynamically weighting several motion models according to their mode probabilities, IMM can handle a broader range of motions than a single model. Nevertheless, when an object performs a maneuver outside the scope of all predefined models, the likelihoods of all modes decrease. The subsequent weighted combination is then formed from models that each approximate the true motion poorly, resulting in increased state-estimation error.

\subsection{Deep-Learning-Based Object Tracking}

Deep-learning-based object-tracking methods frequently predict the next position with a neural network and then use the prediction for data association\supercite{meinhardt2022,wang2025lamotr}. Recurrent networks such as LSTMs were initially used as predictors that replace the motion model of a Kalman filter\supercite{cheng2024}. More recently, Transformer-based predictors have attracted attention because self-attention can capture longer-range temporal dependencies more effectively than recurrent models\supercite{kim2024transformer}. However, these predictors output a future position without explicitly providing how reliable that prediction is. Consequently, when a prediction deviates substantially from the actual observation, no intrinsic uncertainty-based mechanism is available to correct it, and the error can accumulate throughout tracking.

\subsection{Existing Combinations of Kalman Filtering and Deep Learning}

Recent work has also combined Kalman filtering with deep learning. Representative examples include a neural network coupled to the Kalman-filter state model\supercite{bai2025} and an LSTM-based error-compensation method for Kalman filtering\supercite{kim2025lstm}. In these approaches, however, deep learning is used primarily as an auxiliary component of the Kalman filter. When the learned prediction becomes inaccurate, there is no independent alternative mode that can replace or compensate for it, which can degrade tracking performance.

\section{PR-IMM-Based Multiple Object Tracking}

This section presents the proposed PR-IMM tracker, which integrates a Transformer-based prediction model and Kalman-filter-based motion models into a single IMM framework. The Transformer predictor learns nonlinear object motion from data, while the CV, CA, and CT models provide physics-based motion priors that can compensate when the learned prediction becomes unreliable. The goal is to achieve robust position estimation even under noisy observations.

\subsection{Overall System Pipeline}

\begin{figure}[!htbp]
    \centering
    \includegraphics[width=\columnwidth]{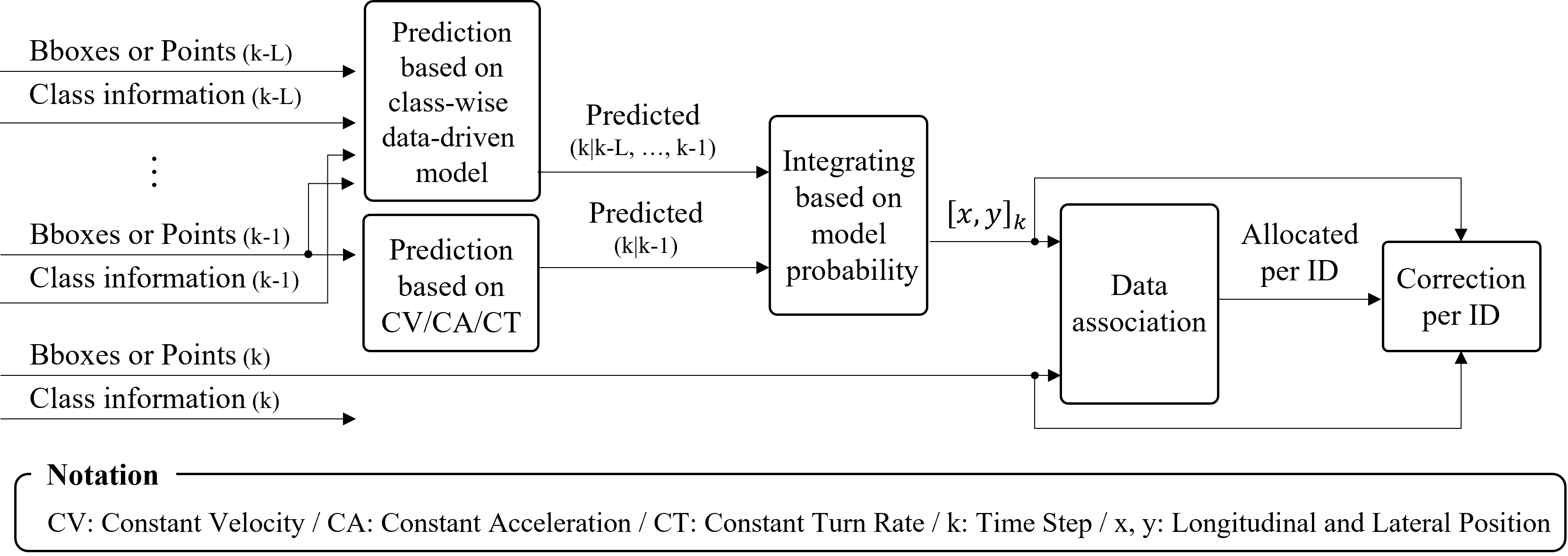}
    \caption{Overall framework of PR-IMM.}
    \label{fig:framework}
\end{figure}

The proposed object-tracking pipeline is shown in Fig.~\ref{fig:framework}. We assume that object detection has already been completed\supercite{kong2025rtnh,pan2024ratrack} and focus on the subsequent tracking stage. Instead of running a detector, we use the point-wise class IDs and track IDs provided by RadarScenes\supercite{schumann2021}. Points with the same track ID are grouped into one object point cloud, and the point values are averaged to obtain an object-level radar measurement vector. At time step $k$, this vector is defined as

\begin{equation}
\mathbf{m}_{k} = \left[x_k,\;y_k,\;v_{r,k},\;\mathrm{RCS}_k\right]^{T},
\label{eq:measurement_full}
\end{equation}
where $x_k$ and $y_k$ denote the longitudinal and lateral distances of the object, respectively, $v_{r,k}$ is radial velocity in the radar coordinate system, and $\mathrm{RCS}_k$ is the Radar Cross Section.

\subsection{Kalman-Filter Prediction with Motion Models}

This subsection describes the Kalman-filter prediction stage based on the CV, CA, and CT motion models. All three models use the two-dimensional object position as the Kalman-filter observation, with

\begin{equation}
\mathbf{z}_{k} = \left[x_k,\;y_k\right]^{T}.
\label{eq:observation}
\end{equation}

Each motion model $i\in\{\mathrm{CV},\mathrm{CA},\mathrm{CT}\}$ uses its own state vector and transition function. The Kalman prediction consists of state prediction

\begin{equation}
\mathbf{x}^{-}_{i,k} = f_i\!\left(\mathbf{x}_{i,k-1}\right),
\label{eq:kf_state_pred}
\end{equation}
and covariance prediction

\begin{equation}
\mathbf{P}^{-}_{i,k}=\mathbf{F}_{i,k}\mathbf{P}_{i,k-1}\mathbf{F}_{i,k}^{T}+\mathbf{Q}_i,
\label{eq:kf_cov_pred}
\end{equation}
where $\mathbf{x}^{-}_{i,k}$ is the prior state predicted by model $i$ from the previous posterior state $\mathbf{x}_{i,k-1}$. The covariance $\mathbf{P}^{-}_{i,k}$ represents the uncertainty of the prior state after incorporating the predefined process-noise covariance $\mathbf{Q}_i$. The superscript $(\cdot)^{-}$ denotes a value before the Kalman correction. For CV and CA, $f_i(\mathbf{x})=\mathbf{F}_i\mathbf{x}$ is linear, whereas CT uses the nonlinear transition in Eqs.~\eqref{eq:ct_x}--\eqref{eq:ct_vy} and propagates covariance using its Jacobian.

\subsubsection{CV Model}

The CV model assumes straight-line motion at constant velocity. Its state vector contains longitudinal and lateral positions and velocities:

\begin{equation}
\mathbf{x}^{-}_{\mathrm{CV},k}=\left[x_k^{-},\;v_{x,k}^{-},\;y_k^{-},\;v_{y,k}^{-}\right]^{T}.
\label{eq:cv_state}
\end{equation}

The initial velocity components $v_x$ and $v_y$ are initialized from the measured radar radial velocity $v_r$ and subsequently refined by Kalman correction. Using the time interval $\Delta t_k$, the state-transition matrix is

\begin{equation}
\mathbf{F}_{\mathrm{CV},k}=
\begin{bmatrix}
1 & \Delta t_k & 0 & 0\\
0 & 1 & 0 & 0\\
0 & 0 & 1 & \Delta t_k\\
0 & 0 & 0 & 1
\end{bmatrix}.
\label{eq:cv_F}
\end{equation}

\subsubsection{CA Model}

The CA model assumes constant acceleration and augments the CV state with acceleration components:

\begin{equation}
\mathbf{x}^{-}_{\mathrm{CA},k}=\left[x_k^{-},\;v_{x,k}^{-},\;a_{x,k}^{-},\;y_k^{-},\;v_{y,k}^{-},\;a_{y,k}^{-}\right]^{T}.
\label{eq:ca_state}
\end{equation}

The initial accelerations are set to zero and are subsequently corrected using measurements. The corresponding transition matrix is

\begin{equation}
\mathbf{F}_{\mathrm{CA},k}=\begin{bmatrix}
1 & \Delta t_k & \tfrac{1}{2}\Delta t_k^2 & 0 & 0 & 0\\
0 & 1 & \Delta t_k & 0 & 0 & 0\\
0 & 0 & 1 & 0 & 0 & 0\\
0 & 0 & 0 & 1 & \Delta t_k & \tfrac{1}{2}\Delta t_k^2\\
0 & 0 & 0 & 0 & 1 & \Delta t_k\\
0 & 0 & 0 & 0 & 0 & 1
\end{bmatrix}.
\label{eq:ca_F}
\end{equation}

\subsubsection{CT Model}

The CT model assumes a constant turn rate $\omega$ and uses

\begin{equation}
\mathbf{x}^{-}_{\mathrm{CT},k}=\left[x_k^{-},\;v_{x,k}^{-},\;y_k^{-},\;v_{y,k}^{-},\;\omega_k^{-}\right]^{T}.
\label{eq:ct_state}
\end{equation}

The initial turn rate is set to zero and is later refined through Kalman correction. Defining $\phi_k=\omega_{k-1}\Delta t_k$, the nonlinear transition for $\omega_{k-1}\neq0$ is

\begin{equation}
 x_k^{-}=x_{k-1}+\frac{\sin\phi_k}{\omega_{k-1}}v_{x,k-1}-\frac{1-\cos\phi_k}{\omega_{k-1}}v_{y,k-1},
\label{eq:ct_x}
\end{equation}

\begin{equation}
 v_{x,k}^{-}=\cos\phi_k\,v_{x,k-1}-\sin\phi_k\,v_{y,k-1},
\label{eq:ct_vx}
\end{equation}

\begin{equation}
 y_k^{-}=y_{k-1}+\frac{1-\cos\phi_k}{\omega_{k-1}}v_{x,k-1}+\frac{\sin\phi_k}{\omega_{k-1}}v_{y,k-1},
\label{eq:ct_y}
\end{equation}

\begin{equation}
 v_{y,k}^{-}=\sin\phi_k\,v_{x,k-1}+\cos\phi_k\,v_{y,k-1}.
\label{eq:ct_vy}
\end{equation}

When $\omega_{k-1}$ is close to zero, a Taylor-series approximation is used to avoid numerical divergence, and covariance propagation is performed using the Jacobian of Eqs.~\eqref{eq:ct_x}--\eqref{eq:ct_vy}.

\subsection{Transformer-Based Prediction}

This subsection describes the Transformer-based prediction model, denoted PR, which predicts the displacement of an object at the next time step from radar observations. The input feature at historical index $l$ is

\begin{equation}
\mathbf{s}_{l}=\left[x_l-x_{k-1},\;y_l-y_{k-1},\;v_{r,l},\;\mathrm{RCS}_l,\;t_l-t_k\right]^{T},
\label{eq:pr_input}
\end{equation}
where $l=k-L,\ldots,k-1$. The PR model uses the previous $L$ radar observations before prediction time $k$, with a maximum sequence length of $L=5$. Each $\mathbf{s}_l$ consists of the relative position with respect to the most recent observation at $k-1$, radial velocity, RCS, and relative time to prediction time $k$.

\begin{figure}[!htbp]
    \centering
    \includegraphics[width=\columnwidth]{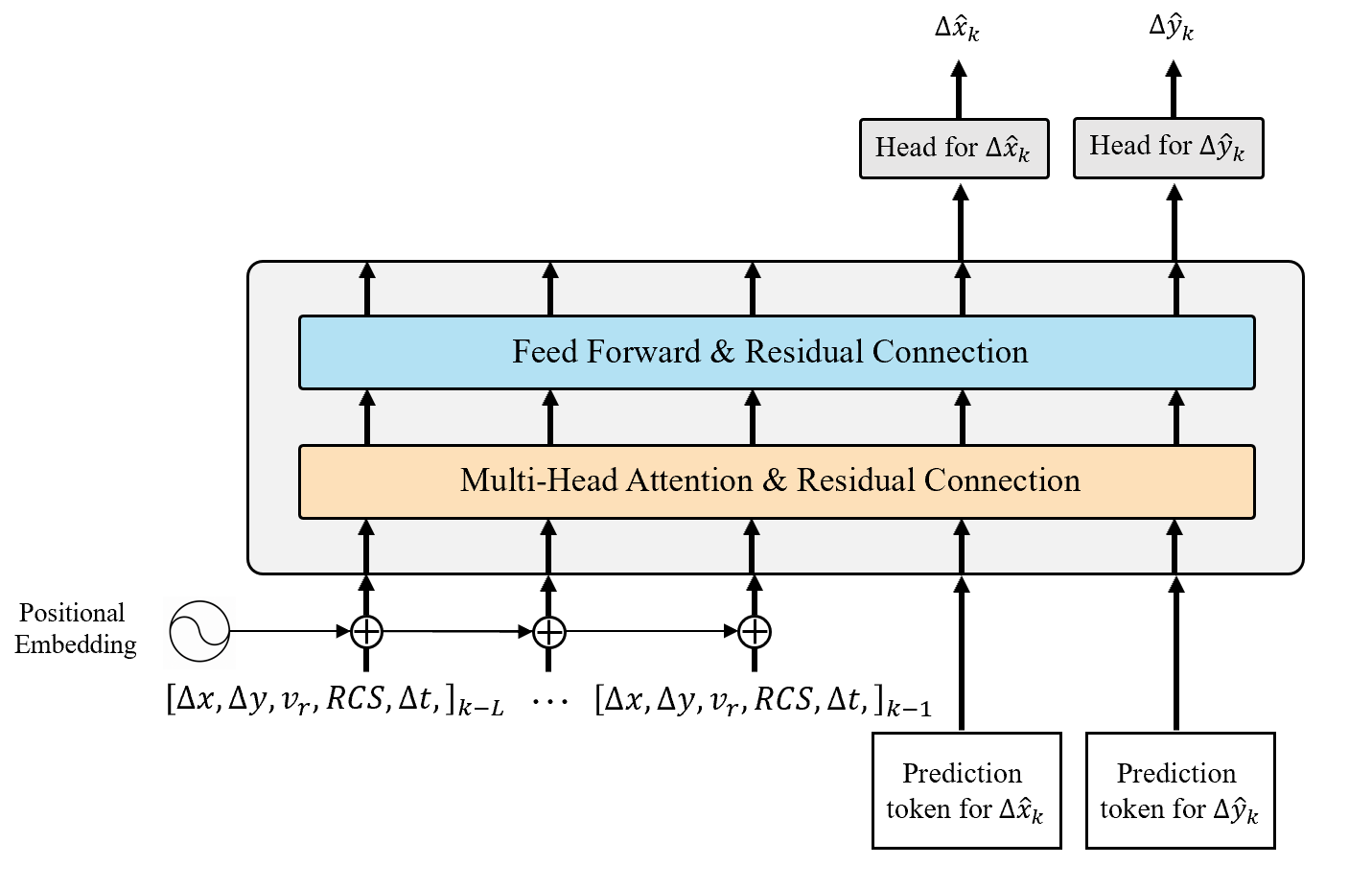}
    \caption{Architecture of the PR model.}
    \label{fig:pr_architecture}
\end{figure}

\subsubsection{PR Model Architecture}

Figure~\ref{fig:pr_architecture} shows the proposed PR architecture. To make the model learn motion patterns rather than absolute scene coordinates, all historical positions are represented relative to the latest observation $\mathbf{z}_{k-1}$. Radial velocity and RCS provide motion and reflection characteristics of the object, and the actual time interval is used to estimate displacement precisely.

Because a Transformer processes an input sequence in parallel, temporal order is not explicitly encoded. Positional embeddings are therefore added to each historical input so that the network receives the temporal ordering of the past trajectory. Two learnable query tokens, $q_x$ and $q_y$, are also appended to the input to predict the $x$- and $y$-direction displacements at the next time step.

The input sequence and query tokens pass through a Transformer encoder composed of Multi-Head Attention and Feed-Forward layers with residual connections. From the final outputs corresponding to the two query tokens, the model predicts
$\Delta\widehat{\mathbf{p}}_{\mathrm{PR},k}=[\Delta\widehat{x}_k,\Delta\widehat{y}_k]^{T}$.

\subsubsection{PR Filter}

The PR prior position at time $k$ is obtained by adding the predicted displacement to the posterior PR position at time $k-1$:

\begin{equation}
\mathbf{p}^{-}_{\mathrm{PR},k}=\mathbf{p}_{\mathrm{PR},k-1}+\Delta\widehat{\mathbf{p}}_{\mathrm{PR},k}.
\label{eq:pr_prior}
\end{equation}

As in the CV, CA, and CT modes, the covariance of the PR prior state represents prediction uncertainty and is defined as

\begin{equation}
\mathbf{P}^{-}_{\mathrm{PR},k}=\mathbf{P}_{\mathrm{PR},k-1}+\mathbf{Q}_{\mathrm{PR}}.
\label{eq:pr_cov}
\end{equation}

\subsection{Prior State Fusion}

This subsection explains how the motion-model predictions in Section III-B and the PR prediction in Section III-C are combined. PR-IMM contains four modes: PR, CV, CA, and CT. Although their complete state vectors have different dimensions, the position component of each prior state is projected through the observation matrix $\mathbf{H}_i$ and used for state fusion:
$\mathbf{p}^{-}_{i,k}=\mathbf{H}_i\mathbf{x}^{-}_{i,k}=[x^{-}_{i,k},y^{-}_{i,k}]^{T}$.

\subsubsection{Input Mixing}

Before the PR, CV, CA, and CT modes predict the next state, each performs input mixing using the previous mode probabilities and mode-transition probabilities. PR-IMM maintains the modes independently, but an object's motion pattern can change over time. To preserve continuity when the dominant mode changes, the posterior positions and position covariances of all modes at the previous time step are combined and used as the initial values for each destination mode. Thus, mode $j$ begins prediction from the mixed position $\overline{\mathbf{p}}_{j,k-1}$, which reflects not only its own previous state but also the posterior states of other modes. The overbar denotes a mode-weighted mean.

Let $\pi_{ij}$ denote the transition probability from mode $i$ to mode $j$, and let $\mu_{i,k-1}$ be the posterior mode probability at the previous time step. The prior probability that mode $j$ is selected at the current time step is

\begin{equation}
 c_{j,k}=\sum_{r=1}^{M}\pi_{rj}\mu_{r,k-1},
\label{eq:prior_mode_prob}
\end{equation}
where $M$ is the total number of modes, namely PR, CV, CA, and CT. The mixing probability that quantifies the contribution of previous mode $i$ to the initial state of current mode $j$ is

\begin{equation}
 \beta_{ij,k-1}=\frac{\pi_{ij}\mu_{i,k-1}}{c_{j,k}}.
\label{eq:mixing_prob}
\end{equation}

Using Eq.~\eqref{eq:mixing_prob}, the mixed initial position and position covariance of mode $j$ are

\begin{equation}
 \overline{\mathbf{p}}_{j,k-1}=\sum_{i=1}^{M}\beta_{ij,k-1}\mathbf{p}_{i,k-1},
\label{eq:mixed_pos}
\end{equation}

\begin{equation}
\begin{split}
 \overline{\mathbf{P}}^{p}_{j,k-1}=\sum_{i=1}^{M}\beta_{ij,k-1}\Big[\mathbf{P}^{p}_{i,k-1}
 +\big(\mathbf{p}_{i,k-1}-\overline{\mathbf{p}}_{j,k-1}\big)\\
 \times\big(\mathbf{p}_{i,k-1}-\overline{\mathbf{p}}_{j,k-1}\big)^{T}\Big].
\end{split}
\label{eq:mixed_pos_cov}
\end{equation}

Equation~\eqref{eq:mixed_pos_cov} includes the dispersion caused by disagreement among the mode positions, so the initial position covariance becomes more conservative when the modes disagree strongly. Because the complete states of the modes have different dimensions and structures, input mixing is performed only for the common position and position covariance.

Each mode independently preserves its non-position state variables. The mixed full state is therefore written as

\begin{equation}
 \overline{\mathbf{x}}_{j,k-1}=\left[\overline{\mathbf{p}}_{j,k-1},\;\mathbf{h}_{j,k-1}\right]^{T},
\label{eq:mixed_state}
\end{equation}
where $\mathbf{h}_{j,k-1}$ contains the non-position states of destination mode $j$ at time $k-1$.

To maintain consistency between the mixed position covariance and the full covariance, the cross-covariance between position and the remaining states and the covariance of the remaining states are updated using the conditional relationship of the original covariance. Before input mixing, the full covariance is partitioned as

\begin{equation}
\mathbf{P}_{j,k-1}=\begin{bmatrix}
\mathbf{P}_{pp,j,k-1} & \mathbf{P}_{ph,j,k-1}\\
\mathbf{P}_{hp,j,k-1} & \mathbf{P}_{hh,j,k-1}
\end{bmatrix},
\label{eq:cov_partition}
\end{equation}
where $\mathbf{P}_{pp}$ is the position covariance, $\mathbf{P}_{ph}=\mathbf{P}_{hp}^{T}$ is the position/non-position cross-covariance, and $\mathbf{P}_{hh}$ is the covariance of the remaining states. We define

\begin{equation}
\mathbf{B}_{j,k-1}=\mathbf{P}_{hp,j,k-1}\left(\mathbf{P}_{pp,j,k-1}\right)^{\dagger},
\label{eq:B}
\end{equation}

\begin{equation}
\mathbf{C}_{j,k-1}=\mathbf{P}_{hh,j,k-1}-\mathbf{B}_{j,k-1}\mathbf{P}_{pp,j,k-1}\mathbf{B}^{T}_{j,k-1},
\label{eq:C}
\end{equation}
and reconstruct the mixed full covariance as

\begin{equation}
\resizebox{0.96\columnwidth}{!}{$\displaystyle
\overline{\mathbf{P}}_{j,k-1}=\begin{bmatrix}
\overline{\mathbf{P}}^{p}_{j,k-1} & \overline{\mathbf{P}}^{p}_{j,k-1}\mathbf{B}^{T}_{j,k-1}\\
\mathbf{B}_{j,k-1}\overline{\mathbf{P}}^{p}_{j,k-1} & \mathbf{C}_{j,k-1}+\mathbf{B}_{j,k-1}\overline{\mathbf{P}}^{p}_{j,k-1}\mathbf{B}^{T}_{j,k-1}
\end{bmatrix}
$}.
\label{eq:reconstructed_cov}
\end{equation}
Here, $(\cdot)^{\dagger}$ denotes the pseudoinverse.

\subsubsection{PR-IMM Prior State Fusion}

Starting from the mixed initial values in Eqs.~\eqref{eq:mixed_state} and \eqref{eq:reconstructed_cov}, each mode independently predicts its prior state. The PR-IMM prior position at time $k$ is then obtained by weighting the predicted position of each mode by its current prior mode probability:

\begin{equation}
\mathbf{p}^{-}_{\mathrm{PR\!-\!IMM},k}=\sum_{i=1}^{M}c_{i,k}\mathbf{p}^{-}_{i,k}.
\label{eq:prior_fusion}
\end{equation}
The fused prior position in Eq.~\eqref{eq:prior_fusion} is used to match tracks and observations in the data-association stage.

\subsection{Data Association}

This subsection describes how the PR-IMM prior state is used to associate existing tracks with new observations. Radar-based multiple-object tracking receives multiple observations at each time step, making correct observation-to-track assignment essential. For track $a$ and observation $j$, we define an association cost using the Euclidean distance between the PR-IMM prior position $\mathbf{p}^{-}_{\mathrm{PR\!-\!IMM},a,k}$ and observation position $\mathbf{z}_{j,k}=[x_{j,k},y_{j,k}]^{T}$ together with their radial-velocity difference:

\begin{equation}
\begin{aligned}
 d_{a,j,k}={}&\left\|\mathbf{z}_{j,k}-\mathbf{p}^{-}_{\mathrm{PR\!-\!IMM},a,k}\right\|_{2}\\
 &+\lambda_v\min\!\left(\left|v_{r,j,k}-v_{r,a}^{\mathrm{last}}\right|,v_{\max}\right).
\end{aligned}
\label{eq:association_cost}
\end{equation}

Here, $v_{r,a}^{\mathrm{last}}$ is the radial velocity of the radar measurement most recently associated with track $a$. This term helps distinguish spatially close objects moving at different velocities. To prevent velocity from dominating the total association cost, the velocity difference is capped by $v_{\max}$ and weighted by $\lambda_v$. We use $\lambda_v=0.1$ and $v_{\max}=5$~m/s. Because RCS varies substantially with object pose, viewing angle, and material composition, it is used only as an input feature to the PR model and is not included in the data-association cost.

For each observation $j$, Eq.~\eqref{eq:association_cost} is used to construct a cost matrix between tracks and observations belonging to the same object class. The Hungarian algorithm then performs a global one-to-one assignment. Associated tracks proceed to the posterior-state-fusion stage using the assigned observation, while observations not associated with any existing track initialize new tracks.

\subsection{Posterior State Fusion}

This subsection explains how each mode is corrected after data association, how the posterior mode probabilities are updated, and how the final position is obtained.

\subsubsection{Mode Probability Update}

Once an observation is assigned to a track, each mode computes the innovation $\boldsymbol{\nu}_{i,k}$ between its prior prediction and the observation and the corresponding innovation covariance $\mathbf{S}_{i,k}$. These quantities are used both for Kalman correction and for computing the mode likelihood and posterior mode probability. For PR, CV, CA, and CT,

\begin{equation}
 \boldsymbol{\nu}_{i,k}=\mathbf{z}_{k}-\mathbf{p}^{-}_{i,k},
\label{eq:innovation}
\end{equation}

\begin{equation}
 \mathbf{S}_{i,k}=\mathbf{H}_i\mathbf{P}^{-}_{i,k}\mathbf{H}_{i}^{T}+\mathbf{R}_i,
\label{eq:innovation_cov}
\end{equation}
where $\mathbf{H}_i$ maps the prior state of mode $i$ to the position observation and $\mathbf{R}_i$ is the radar measurement-noise covariance. For the PR mode, $\mathbf{Q}_{\mathrm{PR}}$ is set smaller than $\mathbf{R}_{\mathrm{PR}}$ so that the learned prior state can be treated as relatively reliable compared with the raw observation.

The Gaussian likelihood measuring how well mode $i$ explains the current observation is

\begin{equation}
 \Lambda_{i,k}=\frac{1}{\sqrt{(2\pi)^2|\mathbf{S}_{i,k}|}}
 \exp\!\left(-\frac{1}{2}\boldsymbol{\nu}_{i,k}^{T}\mathbf{S}_{i,k}^{-1}\boldsymbol{\nu}_{i,k}\right).
\label{eq:likelihood}
\end{equation}

Using this likelihood, the mode probability is updated by

\begin{equation}
 \mu_{i,k}=\frac{\Lambda_{i,k}^{\tau}c_{i,k}}{\sum_{j=1}^{M}\Lambda_{j,k}^{\tau}c_{j,k}},
\label{eq:mode_update}
\end{equation}
where $\tau$ controls the influence of the likelihood. We use $\tau=0.1$.

\subsubsection{Correction}

Using the PR-IMM prior state and the associated observation $\mathbf{z}_k$, each mode updates its posterior state $\mathbf{x}_{i,k}$ and covariance $\mathbf{P}_{i,k}$. The Kalman gain is

\begin{equation}
 \mathbf{K}_{i,k}=\mathbf{P}^{-}_{i,k}\mathbf{H}_{i}^{T}\mathbf{S}_{i,k}^{-1},
\label{eq:kalman_gain}
\end{equation}
followed by the posterior-state update

\begin{equation}
 \mathbf{x}_{i,k}=\mathbf{x}^{-}_{i,k}+\mathbf{K}_{i,k}\boldsymbol{\nu}_{i,k},
\label{eq:posterior_state}
\end{equation}
and the Joseph-form posterior-covariance update

\begin{equation}
\begin{split}
 \mathbf{P}_{i,k}={}&\left(\mathbf{I}-\mathbf{K}_{i,k}\mathbf{H}_i\right)\mathbf{P}^{-}_{i,k}
 \left(\mathbf{I}-\mathbf{K}_{i,k}\mathbf{H}_i\right)^{T}\\
 &+\mathbf{K}_{i,k}\mathbf{R}_i\mathbf{K}_{i,k}^{T}.
\end{split}
\label{eq:posterior_cov}
\end{equation}

Although the observation vector contains only position, the Kalman gain in Eq.~\eqref{eq:kalman_gain} is computed from each mode's full prior covariance. Therefore, the position innovation also updates velocity, acceleration, and turn-rate states through their cross-covariances with position.

\subsubsection{PR-IMM Posterior State Fusion}

The final PR-IMM position is obtained by weighting the posterior position of every mode by the posterior mode probability from Eq.~\eqref{eq:mode_update}:

\begin{equation}
 \mathbf{p}_{\mathrm{PR\!-\!IMM},k}=\sum_{i=1}^{M}\mu_{i,k}\mathbf{H}_i\mathbf{x}_{i,k}.
\label{eq:posterior_fusion}
\end{equation}
Thus, the PR, CV, CA, and CT position estimates are dynamically combined according to the motion situation at each time step.

\section{Experiments}

This section quantitatively and qualitatively evaluates the proposed PR-IMM tracker and analyzes the contribution of its components.

\subsection{Experimental Setup}

\subsubsection{Dataset}

We train the PR model and evaluate the tracker on RadarScenes\supercite{schumann2021}, a large-scale radar point-cloud dataset collected in autonomous-driving environments. RadarScenes provides point-wise segmentation labels and track IDs for five object classes: Car, Large Car, Pedestrian, Pedestrian Group, and Two Wheeler. We merge Car and Large Car into a Car category and Pedestrian and Pedestrian Group into a Pedestrian category, and evaluate only these Car and Pedestrian categories.

RadarScenes contains 158 sequences and is split by sequence for training and evaluation. The validation set uses sequences 6, 42, 58, 85, 99, and 122. The evaluation set uses sequences 1, 2, 3, 4, 16, 33, 38, 41, 47, 49, 57, 71, 81, 98, 105, 112, 113, 120, 123, 125, 131, and 146. The remaining 130 sequences form the training set. Table~\ref{tab:dataset} summarizes the split and class composition.

\begin{table}[!htbp]
    \centering
    \caption{Dataset split and class composition.}
    \label{tab:dataset}
    \scriptsize
    \renewcommand{\arraystretch}{1.08}
    \setlength{\tabcolsep}{3.0pt}
    \resizebox{\columnwidth}{!}{%
    \begin{tabular}{lrrrr}
        \toprule
        \textbf{Split} & \textbf{Sequences} & \textbf{Frames} & \textbf{Car object frames} & \textbf{Pedestrian object frames}\\
        \midrule
        Train      & 130 & 32,555 & 49,349 & 60,216\\
        Validation & 6   & 1,291  & 1,591  & 1,744\\
        Evaluation & 22  & 4,784  & 11,307 & 6,559\\
        \bottomrule
    \end{tabular}}
\end{table}

\subsubsection{Compared Methods}

To evaluate PR-IMM, we quantitatively compare six methods in total. The single-motion-model baselines are CV, CA, and CT Kalman filters. The conventional IMM (CV+CA+CT) serves as the multi-model baseline. We also include the standalone Transformer predictor PR to contrast the limitations of purely physics-based and purely data-driven approaches with the proposed hybrid framework.

\subsubsection{Evaluation Metrics}

Position-prediction performance is evaluated using Root Mean Squared Error (RMSE), while identity preservation in multiple-object tracking is evaluated using ID switches (IDSW) and IDF1.

The longitudinal error $\mathrm{RMSE}_x$, lateral error $\mathrm{RMSE}_y$, and planar distance error $\mathrm{RMSE}_{xy}$ are

\begin{equation}
 \mathrm{RMSE}_x=\sqrt{\frac{1}{N}\sum_{k=1}^{N}\left(x_k-\widehat{x}_k\right)^2},
\label{eq:rmse_x}
\end{equation}

\begin{equation}
 \mathrm{RMSE}_y=\sqrt{\frac{1}{N}\sum_{k=1}^{N}\left(y_k-\widehat{y}_k\right)^2},
\label{eq:rmse_y}
\end{equation}

\begin{equation}
 \mathrm{RMSE}_{xy}=\sqrt{\frac{1}{N}\sum_{k=1}^{N}\left[\left(x_k-\widehat{x}_k\right)^2+\left(y_k-\widehat{y}_k\right)^2\right]},
\label{eq:rmse_xy}
\end{equation}
where $N$ is the total number of samples.

IDSW is incremented by one when the estimated track ID associated with a ground-truth object $g$ at time $k$ differs from the estimated track ID most recently associated with that object.

To compute IDF1, an overlap matrix is first formed for each sequence from the number of object frames in which a ground-truth track ID and an estimated track ID are simultaneously matched. The Hungarian algorithm establishes a one-to-one assignment between the two ID sets that maximizes the total overlap. The overlap sum of the assigned ID pairs is defined as IDTP. Estimated object frames not included in IDTP are IDFP, while ground-truth object frames not included in IDTP are IDFN. IDF1 is then

\begin{equation}
 \mathrm{IDF1}=\frac{2\,\mathrm{IDTP}}{2\,\mathrm{IDTP}+\mathrm{IDFP}+\mathrm{IDFN}}.
\label{eq:idf1}
\end{equation}

The IDTP, IDFP, and IDFN values are summed over all evaluation sequences before the final IDF1 is computed. Lower IDSW and higher IDF1 indicate better identity-preservation performance.

\subsection{Quantitative Performance Comparison}

Table~\ref{tab:prior_rmse} compares one-step prior prediction accuracy. PR-IMM achieves the lowest error among all methods. This result indicates that the general physics-based motion models compensate when the data-driven PR prediction becomes unstable, while the mode probabilities automatically assign greater weight to the mode that best explains the current motion.

\begin{table}[!htbp]
    \centering
    \caption{One-step prior prediction RMSE.}
    \label{tab:prior_rmse}
    \scriptsize
    \renewcommand{\arraystretch}{1.08}
    \resizebox{\columnwidth}{!}{%
    \begin{tabular}{lccc}
        \toprule
        \textbf{Model} & $\mathbf{RMSE_x\downarrow}$ [m] & $\mathbf{RMSE_y\downarrow}$ [m] & $\mathbf{RMSE_{xy}\downarrow}$ [m]\\
        \midrule
        CV & 4.884 & 3.801 & 6.189\\
        CA & 2.241 & 2.068 & 3.050\\
        CT & 4.872 & 3.545 & 6.025\\
        PR & 0.971 & 0.822 & 1.272\\
        IMM (CV+CA+CT) & 1.827 & 1.685 & 2.485\\
        \textbf{PR-IMM (Ours)} & \textbf{0.799} & \textbf{0.698} & \textbf{1.062}\\
        \bottomrule
    \end{tabular}}
\end{table}

RadarScenes does not provide labels for motion scenarios, so scenario categories are derived from the previous ten ground-truth positions, timestamps, and the object's heading. Objects that satisfy none of the scenario conditions are excluded from scenario-wise evaluation. As shown in Table~\ref{tab:scenario_rmse}, PR-IMM obtains the lowest $\mathrm{RMSE}_{xy}$ in all four motion scenarios. In particular, it outperforms the standalone PR model even in the Nonlinear scenario, which is difficult to explain with a single motion assumption. This demonstrates that combining data-driven prediction with physics-based motion models remains effective for complex motion.

\begin{table}[!t]
    \centering
    \caption{Scenario-wise prior prediction RMSE.}
    \label{tab:scenario_rmse}
    \scriptsize
    \setlength{\tabcolsep}{2.2pt}
    \renewcommand{\arraystretch}{1.05}

    \begin{tabular}{lcccccc}
        \toprule
        \textbf{Scenario} &
        \textbf{CV} &
        \textbf{CA} &
        \textbf{CT} &
        \textbf{PR} &
        \textbf{IMM} &
        \textbf{Ours} \\
        \midrule
        Straight & 4.007 & 1.308 & 5.119 & 0.633 & 1.038 & \textbf{0.584} \\
        Accel./Decel. & 8.387 & 2.323 & 8.694 & 1.029 & 2.099 & \textbf{0.984} \\
        Left/Right Turn & 7.055 & 2.999 & 6.781 & 1.024 & 2.291 & \textbf{0.904} \\
        Nonlinear & 7.433 & 3.896 & 7.164 & 1.270 & 3.143 & \textbf{1.106} \\
        \bottomrule
    \end{tabular}
\end{table}

\FloatBarrier

Table~\ref{tab:overall_mot} reports overall IDSW and IDF1.
Compared with the standalone PR model, PR-IMM reduces IDSW
from 915 to 519 and increases IDF1 from 0.8795 to 0.9020.
This shows that the proposed method improves not only position
RMSE but also the ability to preserve track IDs when multiple
objects are close to one another.

\begin{table}[!htbp]
    \centering
    \caption{Overall multiple-object-tracking performance (GT object frames = 17,866).}
    \label{tab:overall_mot}
    \scriptsize
    \renewcommand{\arraystretch}{1.08}
    \begin{tabular}{lcc}
        \toprule
        \textbf{Model} & $\mathbf{IDSW\downarrow}$ & $\mathbf{IDF1\uparrow}$\\
        \midrule
        CV & 954 & 0.7583\\
        CA & 700 & 0.8075\\
        CT & 901 & 0.7571\\
        PR & 915 & 0.8795\\
        IMM (CV+CA+CT) & 695 & 0.8228\\
        \textbf{PR-IMM (Ours)} & \textbf{519} & \textbf{0.9020}\\
        \bottomrule
    \end{tabular}
\end{table}

\begin{figure*}[!t]
    \centering
    \includegraphics[width=0.96\textwidth]{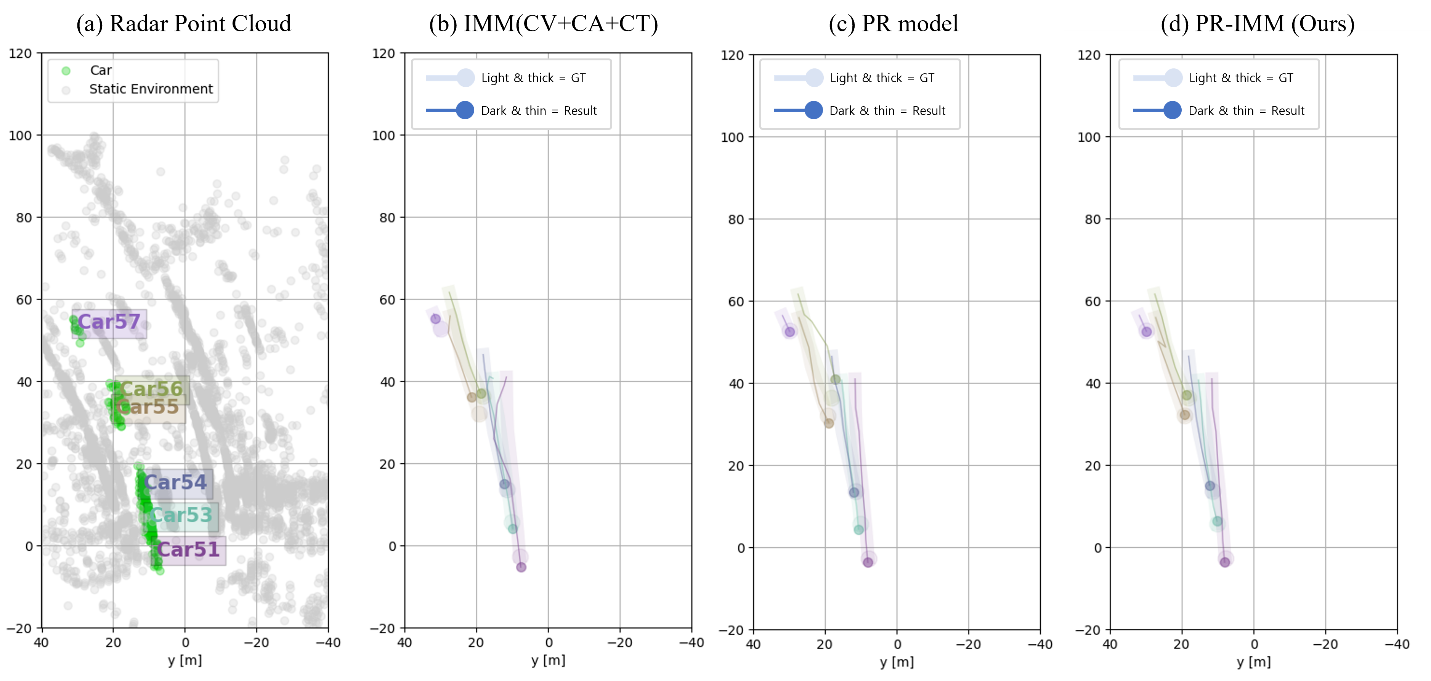}
    \caption{Qualitative tracking results on the RadarScenes dataset. (a) Radar point cloud with annotated vehicle points. (b) CV/CA/CT-based IMM result. (c) PR model result. (d) Proposed PR-IMM result. In (b)--(d), light thick lines denote ground-truth trajectories, whereas dark thin lines denote estimated trajectories.}
    \label{fig:tracking_results}
\end{figure*}

To analyze tracking performance under more difficult conditions, we additionally construct scenarios in which data association becomes ambiguous because objects are spatially close or their paths cross. Table~\ref{tab:scenario_mot} summarizes these results. Close Proximity denotes frames in which the nearest object is within 4~m. Crossing Motion denotes cases in which two objects are within 8~m and their heading difference is between $45^{\circ}$ and $135^{\circ}$. Missed-detection scenarios are generated by removing 10\% or 20\% of the measurements in each sequence using the same random seed. PR-IMM achieves the lowest IDSW and the highest IDF1 in every scenario. Thus, it remains robust even when data association is more ambiguous or observations are partially missing.

\begin{table}[!htbp]
    \centering
    \caption{Scenario-wise multiple-object-tracking performance.}
    \label{tab:scenario_mot}
    \scriptsize
    \renewcommand{\arraystretch}{1.05}
    \setlength{\tabcolsep}{2.7pt}
    \resizebox{\columnwidth}{!}{%
    \begin{tabular}{llrrr}
        \toprule
        \textbf{Scenario (GT object frames)} & \textbf{Metric} & \textbf{PR} & \textbf{IMM} & \textbf{PR-IMM (Ours)}\\
        \midrule
        \multirow{2}{*}{Close Proximity (1,284)} & IDSW$\downarrow$ & 54 & 43 & \textbf{34}\\
        & IDF1$\uparrow$ & 0.8769 & 0.8100 & \textbf{0.8941}\\
        \multirow{2}{*}{Crossing Motion (687)} & IDSW$\downarrow$ & 20 & 21 & \textbf{11}\\
        & IDF1$\uparrow$ & 0.9127 & 0.8486 & \textbf{0.9258}\\
        \multirow{2}{*}{10\% Miss (17,866)} & IDSW$\downarrow$ & 1,245 & 836 & \textbf{711}\\
        & IDF1$\uparrow$ & 0.7316 & 0.7185 & \textbf{0.7610}\\
        \multirow{2}{*}{20\% Miss (17,866)} & IDSW$\downarrow$ & 1,952 & 1,197 & \textbf{1,146}\\
        & IDF1$\uparrow$ & 0.5476 & 0.5629 & \textbf{0.5849}\\
        \bottomrule
    \end{tabular}}
\end{table}

\subsection{Qualitative Analysis}

This subsection qualitatively analyzes PR-IMM through mode-probability statistics and trajectory visualizations.

\subsubsection{Mode Probability Visualization}

\begin{figure}[!htbp]
    \centering
    \includegraphics[width=\columnwidth]{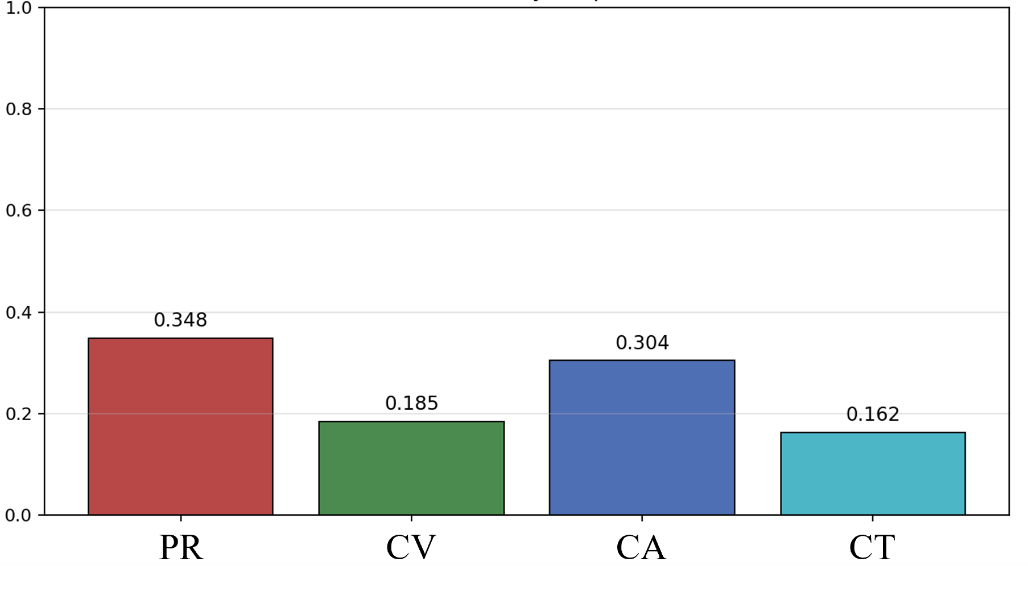}
    \caption{Mean mode probability of PR-IMM.}
    \label{fig:mode_probability}
\end{figure}

Figure~\ref{fig:mode_probability} shows the average probability of each PR-IMM mode over the entire test dataset. PR has the highest mean probability at 0.348, followed by CA at 0.304, CV at 0.185, and CT at 0.162. PR receives the largest share because the Transformer predictor frequently provides a prior state close to the observation based on the recent trajectory and therefore tends to obtain a relatively high likelihood.

However, the PR probability is only 0.348 and is therefore not dominant by itself; more than half of the total probability is distributed among CV, CA, and CT. The CA mode has a higher average probability than CV and CT, suggesting that the dataset contains a substantial number of accelerating and decelerating intervals in which CA complements the PR prediction. These results support the reason why PR-IMM outperforms PR alone: tracking does not rely on a single mode, but rather combines the data-driven predictor and physics-based motion models in a complementary manner.

\subsubsection{Object Trajectory Visualization}

Figure~\ref{fig:tracking_results} presents qualitative object-tracking results. The conventional IMM (CV+CA+CT) shows the largest deviations of multi-object trajectories from the ground truth, with particularly noticeable dispersion where objects are close to one another. The PR model follows the ground truth more closely than the conventional IMM, but its estimated positions still fluctuate in some intervals. In contrast, PR-IMM produces trajectories that most closely match the ground truth throughout the sequence and tracks each object stably even when multiple objects are adjacent. These observations are consistent with the RMSE results and demonstrate that PR-IMM robustly combines data-driven and physics-based prior states in both quantitative and qualitative evaluations.

\section{Conclusion}

This paper proposed a PR-IMM tracker for radar-point-cloud-based multiple-object tracking. The PR model converts radar observations into relative displacements with respect to the most recent observation and uses a Transformer encoder to predict the next displacement, allowing nonlinear object motion to be learned directly from data without a fixed motion assumption. The PR model is then integrated as an independent IMM mode together with CV, CA, and CT. Because the four modes have different state dimensions, input mixing and state fusion are performed using the two-dimensional position and position covariance shared by all modes, enabling data-driven and physics-based prior states to be dynamically combined according to the current motion situation.

Experiments show that PR-IMM achieves the best performance among the CV, CA, and CT single-motion models, the conventional IMM (CV+CA+CT), and the standalone PR model. Analysis of mode probabilities and tracking trajectories further confirms the complementary operation of the PR mode and the physics-based modes. These results demonstrate that integrating a learned state predictor into the probabilistic IMM framework can improve both position-estimation accuracy and identity-preserving multiple-object tracking under diverse motion and observation conditions.

\section*{Acknowledgment}

This work was supported by the Ministry of SMEs and Startups, Republic of Korea, in 2026 under Grant RS-2025-24535910.

\bibliographystyle{IEEEtran}
\bibliography{PR_IMM}

@article{han2025,
  author  = {Geonhee Han and Seok-Cheol Kee},
  title   = {A Hybrid Approach Multi-Object Tracking with Optical Flow and Kalman Filter},
  journal = {Transactions of the Korean Society of Automotive Engineers},
  year    = {2025},
  volume  = {33},
  number  = {9},
  pages   = {755--762},
  doi     = {10.7467/KSAE.2025.33.9.755}
}

@article{yoon2024,
  author  = {Dohyun Yoon and Keonwoo Jang and Jongjin Won and Yeonsik Kang},
  title   = {A Multi-task Deep Learning Method for Road Environment Recognition and Object Tracking},
  journal = {Transactions of the Korean Society of Automotive Engineers},
  year    = {2024},
  volume  = {32},
  number  = {1},
  pages   = {77--82},
  doi     = {10.7467/KSAE.2024.32.1.77}
}

@article{jung2025,
  author  = {Hee-Yang Jung and Dong-Hee Paek and Seung-Hyun Kong},
  title   = {Open-Source Autonomous Driving Software Platforms: Comparison of {Autoware} and {Apollo}},
  journal = {arXiv preprint arXiv:2501.18942},
  year    = {2025}
}

@inproceedings{cho2019,
  author    = {Sang Jae Cho and Bo Seong Kim and Tae Seon Kim and Seung-Hyun Kong},
  title     = {Enhancing {GNSS} Performance and Detection of Road Crossing in Urban Area Using Deep Learning},
  booktitle = {2019 IEEE Intelligent Transportation Systems Conference (ITSC)},
  year      = {2019},
  pages     = {2115--2120}
}

@article{kong2013,
  author  = {Seung-Hyun Kong and Binhee Kim},
  title   = {Two-Dimensional Compressed Correlator for Fast {PN} Code Acquisition},
  journal = {IEEE Transactions on Wireless Communications},
  year    = {2013},
  volume  = {12},
  number  = {11},
  pages   = {5859--5867},
  doi     = {10.1109/TWC.2013.092313.130407}
}

@article{kong2014,
  author  = {Seung-Hyun Kong},
  title   = {Fast Multi-Satellite {ML} Acquisition for {A-GPS}},
  journal = {IEEE Transactions on Wireless Communications},
  year    = {2014},
  volume  = {13},
  number  = {9},
  pages   = {4935--4946}
}

@article{jan2013,
  author  = {Shau-Shiun Jan and Yu-Chun Kao},
  title   = {Radar Tracking with an Interacting Multiple Model and Probabilistic Data Association Filter for Civil Aviation Applications},
  journal = {Sensors},
  year    = {2013},
  volume  = {13},
  number  = {5},
  pages   = {6636--6650},
  doi     = {10.3390/s130506636}
}

@inproceedings{manjunath2018,
  author    = {A. Manjunath and Y. Liu and B. Henriques and A. Engstle},
  title     = {Radar Based Object Detection and Tracking for Autonomous Driving},
  booktitle = {2018 IEEE MTT-S International Conference on Microwaves for Intelligent Mobility (ICMIM)},
  year      = {2018},
  pages     = {1--4}
}

@article{blom1988,
  author  = {H. A. P. Blom and Y. Bar-Shalom},
  title   = {The Interacting Multiple Model Algorithm for Systems with Markovian Switching Coefficients},
  journal = {IEEE Transactions on Automatic Control},
  year    = {1988},
  volume  = {33},
  number  = {8},
  pages   = {780--783}
}

@inproceedings{lv2024,
  author    = {W. Lv and Y. Huang and N. Zhang and R. S. Lin and M. Han and D. Zeng},
  title     = {DiffMOT: A Real-time Diffusion-based Multiple Object Tracker with Non-linear Prediction},
  booktitle = {2024 IEEE/CVF Conference on Computer Vision and Pattern Recognition (CVPR)},
  year      = {2024},
  pages     = {19321--19330}
}

@inproceedings{kim2025bayes,
  author    = {Dong-In Kim and Dong-Hee Paek and Seung-Hyun Song and Seung-Hyun Kong},
  title     = {Bayesian Approximation-Based Trajectory Prediction and Tracking with 4D Radar},
  booktitle = {2025 IEEE Intelligent Vehicles Symposium (IV)},
  year      = {2025},
  pages     = {1380--1385}
}

@inproceedings{vaswani2017,
  author    = {Ashish Vaswani and Noam Shazeer and Niki Parmar and Jakob Uszkoreit and Llion Jones and Aidan N. Gomez and Lukasz Kaiser and Illia Polosukhin},
  title     = {Attention Is All You Need},
  booktitle = {Advances in Neural Information Processing Systems},
  year      = {2017},
  volume    = {30},
  pages     = {5998--6008}
}

@article{kim2024transformer,
  author  = {M. Kim and B. I. Kwak and J. U. Hou and T. Kim},
  title   = {Robust Long-Term Vehicle Trajectory Prediction Using Link Projection and a Situation-Aware Transformer},
  journal = {Sensors},
  year    = {2024},
  volume  = {24},
  number  = {8},
  pages   = {2398},
  doi     = {10.3390/s24082398}
}

@article{tang2024,
  author  = {X. Tang and X. Cheng and N. Xu},
  title   = {A Robust Multiobject Tracking Method Based on 4-D Millimeter-Wave Radar and Monocular Vision Fusion},
  journal = {IEEE Sensors Journal},
  year    = {2024},
  volume  = {24},
  number  = {22},
  pages   = {37764--37774}
}

@inproceedings{schumann2021,
  author    = {O. Schumann and M. Hahn and N. Scheiner and F. Weishaupt and J. F. Tilly and J. Dickmann and C. W{\"o}hler},
  title     = {RadarScenes: A Real-World Radar Point Cloud Data Set for Automotive Applications},
  booktitle = {2021 IEEE 24th International Conference on Information Fusion (FUSION)},
  year      = {2021},
  pages     = {1--8}
}

@inproceedings{bewley2016,
  author    = {Alex Bewley and Zongyuan Ge and Lionel Ott and Fabio Ramos and Ben Upcroft},
  title     = {Simple Online and Realtime Tracking},
  booktitle = {2016 IEEE International Conference on Image Processing (ICIP)},
  year      = {2016},
  pages     = {3464--3468},
  doi       = {10.1109/ICIP.2016.7533003}
}

@article{li2019,
  author  = {Y. Li and C. Liang and M. Lu and X. Hu and Y. Wang},
  title   = {Cascaded Kalman Filter for Target Tracking in Automotive Radar},
  journal = {The Journal of Engineering},
  year    = {2019},
  volume  = {2019},
  number  = {19},
  pages   = {6264--6267}
}

@inproceedings{li2024immekf,
  author    = {Q. Li and Z. Zhang and G. He and X. Hu and X. Kang},
  title     = {3D Multi-Object Tracking for Autonomous Driving Based on {IMM-EKF} and Re-Identification},
  booktitle = {2024 IEEE International Conference on Unmanned Systems (ICUS)},
  year      = {2024},
  pages     = {1197--1202}
}

@inproceedings{liu2025immmot,
  author    = {Xiaohong Liu and Xulong Zhao and Gang Liu and Zili Wu and Tao Wang and Lei Meng and Yuhan Wang},
  title     = {{IMM-MOT}: A Novel 3D Multi-object Tracking Framework with Interacting Multiple Model Filter},
  booktitle = {2025 IEEE/RSJ International Conference on Intelligent Robots and Systems (IROS)},
  year      = {2025},
  pages     = {20540--20547}
}

@inproceedings{meinhardt2022,
  author    = {Tim Meinhardt and Alexander Kirillov and Laura Leal-Taixe and Christoph Feichtenhofer},
  title     = {TrackFormer: Multi-Object Tracking With Transformers},
  booktitle = {2022 IEEE/CVF Conference on Computer Vision and Pattern Recognition (CVPR)},
  year      = {2022},
  pages     = {8844--8854}
}

@inproceedings{wang2025lamotr,
  author    = {Peng Wang and Yongcai Wang and Hualong Cao and Wang Chen and Deying Li},
  title     = {{LA-MOTR}: End-to-End Multi-Object Tracking by Learnable Association},
  booktitle = {2025 IEEE/CVF International Conference on Computer Vision (ICCV)},
  year      = {2025},
  pages     = {12438--12448},
  doi       = {10.1109/ICCV51701.2025.01156}
}

@article{cheng2024,
  author  = {L. Cheng and A. Sengupta and S. Cao},
  title   = {Deep Learning-Based Robust Multi-Object Tracking via Fusion of mmWave Radar and Camera Sensors},
  journal = {IEEE Transactions on Intelligent Transportation Systems},
  year    = {2024},
  volume  = {25},
  number  = {11},
  pages   = {17218--17233}
}

@article{bai2025,
  author  = {Y. Bai and B. Yan and W. Dong and X. Jin and T. Su and H. Ma},
  title   = {A Novel Adaptive State Estimation Model: Kalman Filter Coupled with Neural Networks},
  journal = {International Journal of Adaptive Control and Signal Processing},
  year    = {2025},
  volume  = {39},
  number  = {5},
  pages   = {914--926},
  doi     = {10.1002/acs.3982}
}

@article{kim2025lstm,
  author  = {Sohyun Kim and Jaehoon Choi and Dayeon Jeong and Seokwon Lee},
  title   = {{LSTM}-Based Kalman Filter Error Compensation for Enhanced Moving Object Tracking},
  journal = {Journal of Institute of Control, Robotics and Systems},
  year    = {2025},
  volume  = {31},
  number  = {1},
  pages   = {68--76},
  doi     = {10.5302/J.ICROS.2025.24.0143}
}

@article{kong2025rtnh,
  author  = {Seung-Hyun Kong and Dong-Hee Paek and Sangyeong Lee},
  title   = {{RTNH+}: Enhanced 4D Radar Object Detection Network Using Two-Level Preprocessing and Vertical Encoding},
  journal = {IEEE Transactions on Intelligent Vehicles},
  year    = {2025},
  volume  = {10},
  number  = {2},
  pages   = {1427--1440},
  doi     = {10.1109/TIV.2024.3428696}
}

@inproceedings{pan2024ratrack,
  author    = {Z. Pan and F. Ding and H. Zhong and C. X. Lu},
  title     = {RaTrack: Moving Object Detection and Tracking with 4D Radar Point Cloud},
  booktitle = {2024 IEEE International Conference on Robotics and Automation (ICRA)},
  year      = {2024},
  pages     = {4480--4487}
}

\end{document}